\documentclass[a4paper, 10pt, conference]{ieeeconf}      

\usepackage[utf8]{inputenc}
\usepackage[T1]{fontenc}
\usepackage{graphics} 
\usepackage{epsfig}   
\usepackage{mathptmx} 
\usepackage{times}    
\usepackage{amsmath}  
\usepackage{amssymb}  
\usepackage{ifthen}
\usepackage{lettrine}
\usepackage{booktabs}
\usepackage{url}
\usepackage{hyperref}
\usepackage{xcolor}
\usepackage{physics}
\usepackage{geometry}
\usepackage{algorithm} 
\usepackage{algpseudocode} 
\graphicspath{ {./pic/} }

\title{\LARGE \bf
Trajectory Prediction \& Path Planning for an Object Intercepting UAV with a Mounted Depth Camera}

\author{Jasper Tan$^{1*}$, Arijit Dasgupta$^{2*}$, Arjun Agrawal$^{3*}$ and Sutthiphong Srigrarom$^{4}$
\thanks{$^{*}$These authors contributed equally.}
\thanks{$^{1}$Jasper Tan is with National University of Singapore, 5A Engineering Drive 1, Singapore 117411 
        {\tt\small jasper.tan@u.nus.edu}}%
\thanks{$^{2}$Arijit Dasgupta is with National University of Singapore, 5A Engineering Drive 1, Singapore 117411 
        {\tt\small arijit.dasgupta@u.nus.edu}}%
\thanks{$^{3}$Arjun Agrawal is with National University of Singapore, 5A Engineering Drive 1, Singapore 117411 
        {\tt\small Arjun.a@u.nus.edu}}%
\thanks{$^{4}$Sutthiphong Srigrarom is with National University of Singapore, 5A Engineering Drive 1, Singapore 117411 
        {\tt\small spot.srigrarom@nus.edu.sg}}%
}

\begin{document}

\maketitle
\thispagestyle{empty}
\pagestyle{empty}

\begin{abstract}
A novel control \& software architecture using ROS C++ is introduced for object interception by a UAV with a mounted depth camera and no external aid. Existing work in trajectory prediction focused on the use of off-board tools like motion capture rooms to intercept thrown objects. The present study designs the UAV architecture to be completely on-board capable of object interception with the use of a depth camera and point cloud processing. The architecture uses an iterative trajectory prediction algorithm for non-propelled objects like a ping-pong ball. A variety of path planning approaches to object interception and their corresponding scenarios are discussed, evaluated \& simulated in Gazebo. The successful simulations exemplify the potential of using the proposed architecture for the on-board autonomy of UAVs intercepting objects.

\(Keywords:\) Depth Camera, Trajectory Prediction, Object Interception, Path Planning, Autonomous Vehicle Systems. 
\end{abstract}

\section{Introduction}


Unmanned Aerial Vehicles (UAVs) have been involved in a great number of diverse real-time applications such as inspection and search \& rescue. From a control perspective, this entailed near-hover operations as detailed in \cite{Indoor_env}. Today UAVs are required to interact independently with their environment and modify their trajectories to adapt to changes in the environment. This requires more dynamic, precise and aggressive maneuvers to be performed as demonstrated in \cite{precise_aggressive_maneuvers_quadrotors} and \cite{quad_flips}. A common implementation of this is found in last-mile delivery UAVs and game-based UAV applications. Game-based UAV applications have generated many exciting yet difficult tasks which demonstrate valuable real-time applications for the scientific community. A key example of this is of juggling a ball \cite{Quad_Ball_Juggle}. Algorithms that generate these trajectories with the use of lines, polynomials and splines while respecting the dynamic constraints of the UAV have been outlined by several authors \cite{Hoffmann2007_line, Cowling2007_polynomial} \& \cite{Bouktir_Spline}. Researchers have also worked on time optimised trajectory planning algorithms where the UAV must attain a set position within a given time frame \cite{Quadrocopters_Real-Time_Trajectory_Generation}.

The present study aims to push the boundary of UAV trajectory planning. One salient disadvantage in the aforementioned papers is that the localisation and trajectory planning functions are aided by a motion capture room. Instead, we propose a different method - to use depth cameras to identify the location of a projectile concerning the UAV \& use odometry to localise the object with the room. This approach brings up a multitude of new problems. It needs a robust method to identify the object \& inform the UAV's autopilot about the location of the object. The approach also needs a good trajectory prediction algorithm so that the UAV can intercept the object.

To serve as a proof of concept, our conceptualisation makes use of a ping-pong ball projectile that the UAV needs to intercept. Intercepting a ball is a visually engaging problem that can easily function as an indication of how successful the system is. The problem of intercepting the ball allows for the opportunity to explore strategies to identify and predict projectile trajectories. The objective of this paper is to conceptualise \& build a control \& architecture software that is integrated with MAVROS and PX4 to intercept a ball projectile. We test our control \& architecture software inside a Gazebo simulation environment as a form of proof-of-concept. To meet the standards of real-world and real-time systems, the architecture is designed such that there is no reliance on any external tools, like cameras fixed in the environment or motion capture tools. All tools used will be present on board the UAV. This report makes the \textbf{assumption} that any UAV using our software has a PX4 flight controller, a companion computer \& a depth camera.


\section{Related Works}

Early research in the field of UAV control revolved around the near-hover condition. Linear controllers were used and tuned using methods described in \cite{PID_linearControl}. In these implementations, trajectories are generated with the help of way-points and limited by low flight velocities to meet the near-hover condition \cite{Indoor_env}.

To better exploit the advantages which make UAVs attractive, more complex control strategies are required. UAVs offer instantaneous agile movement in three-dimensional space. They often also provide a high thrust to weight ratio enabling UAVs to carry out fast transnational dynamics. Interesting implementations of controllers which exploit these capabilities have been presented in \cite{quad_flips} through the performance of flips and in \cite{augugliaro2013dance} through a coordinated dance routine of UAVs. Several algorithms which generate trajectories in space from path primitives - splines \cite{Bouktir_Spline}, polynomials \cite{Cowling2007_polynomial} and lines \cite{Hoffmann2007_line} - based on the differential flatness of UAV dynamics have been developed. These algorithms operate on a two-step process. First, a sample-based trajectory planning algorithm is used to create a path between the UAV and the setpoint. Then, the generated path is parameterised in time to enforce the UAV's dynamic constraints - hence ensuring that the generated path is feasible.

There are also algorithms that exploit the UAV's differential flatness property to minimise a derivative of the position trajectory such as the snap. Two implementations of these kinds of algorithms are the minimum snap trajectory generation \cite{Minimum_snap_trajectory_generation} and a method which aims to minimise the weighted sum of the trajectory derivatives \cite{Polynomial_Trajectory_Differential}. Another algorithm presented in \cite{LBMPC}, focuses on the real-time implementation of trajectory planning with the help of learning-based model predictive control. The algorithm is robust and used to intercept a ball thrown with an unknown trajectory within a motion capture room. 
 
The next group of algorithms aimed to provide a time-optimal solution to the trajectory problem by considering the non-linear dynamics of the UAV. Solutions use well established control methods such as the Pontryagin’s minimum principle \cite{time-optimal_Pontryagin} and the numerical optimal control \cite{Time-optimal_quadrotor_flight}. However, these approaches relied on perception capabilities external to the UAV. The present study recognises the need to use on-board capabilities for autonomous perception and trajectory prediction. Note that the term UAV is used interchangeably in the following sections.

\section{Methodology}

\subsection{Proposed Architecture}

Our proposed architecture can be split into three sequential stages: object detection, trajectory prediction \& path planning. The Robot Operating System (ROS) was used as the software framework for the proposed architecture. As the flight time of the object is expected to be very short (\(<1s\)), all code is written in C++ language, using minimal libraries to reduce latency and maximise frames for trajectory prediction. The code for the ROS Workspace is made publicly available in \href{https://github.com/arijitnoobstar/UAVProjectileCatcher}{\color{red}{GitHub}}.

\subsubsection{Object Detection}

The present study only considers a simple colour thresholding \& fringe point filtering method to detect a unicolour object like a ping pong ball. This is because the present step is not the emphasis of the architecture. Developers may use data-driven CNN-based object detection algorithms like YOLOv3\cite{farhadi2018yolov3} \& ResNet\cite{he2016deep} to filter the points that represent the object. Using a depth camera, an RGBXYZ point cloud can be constructed around the UAV by aligning the depth map to the RGB image using the Point Cloud Library (PCL) \cite{Rusu_ICRA2011_PCL}. Colour thresholding is implemented to detect the points that represent the ball and all fringe points one standard deviation from the mean location are filtered. Finally, the centroid location of the points is determined as the location of the ball, ($X_{object}, Y_{object}, Z_{object}$), in 3D space. Such processing is done in almost instantly, and no drop in framerate was observed.

\subsubsection{Trajectory Prediction}
\label{trajpred}

Predicting the trajectory of the object is the most crucial step of the architecture.

\subsubsection*{Step 1: Find the Velocity}

The velocity ($U_{object}, V_{object}, W_{object}$) can be easily measured by observing how $X_{object}, Y_{object}, Z_{object}$ changes with time, followed by a least-squares linear regression to take the gradient as velocity. Equation \ref{linearreg} shows the calculation of the speed in the $i$ direction to get $i_{object}$. With a small time-step, we make a constant velocity assumption that between successive measurements of $X_{object}, Y_{object}, Z_{object}$. As our system is real-time, the prediction algorithm uses a queue data structure to implement the regression where a fixed-queue size allows for a moving-average calculation of the velocity with time. The queue size was tuned to 5 to balance the constant velocity assumption and minimise random error.

\begin{equation}
    \label{linearreg}
    i_{object} = \frac{n\sum ti_{object} - \sum i \sum t}{n\sum t^2 - (\sum t)^2}
\end{equation}

\subsubsection*{Step 2: Determine the Acceleration}

\begin{figure}
    \centering
    \includegraphics[width = 0.3\linewidth]{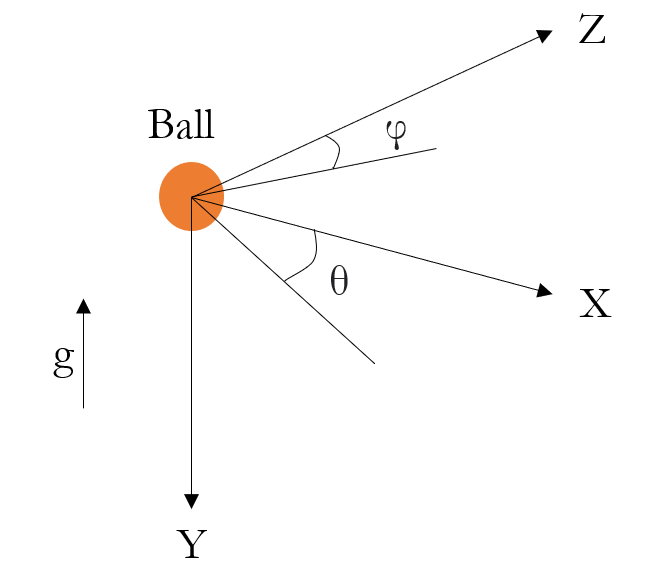}
    \caption{World Coordinate System of the ball}
    \label{fig:worldcoordinate}
\end{figure}

Figure \ref{fig:worldcoordinate} illustrates the world coordinate frame of the ball. To determine the acceleration of the ball, its weight and drag of must be resolved.  Assuming standard ping pong ball characteristics and room temperature conditions, the Reynolds number is estimated as $\frac{VD}{\nu} \cong 2700$. Based on the equation \ref{cd} by \cite{morrison2013data}, we arrive at a $C_D$ of $\cong 0.420$. For objects of non-spherical shapes, a separate method of $C_D$ estimation can be used\cite{chhabra1999drag}. The velocity magnitude, $\abs{V} = \sqrt{U_{object}^2 + V_{object}^2 + W_{object}^2}$, is also determined. The drag, $D_r$, of the ball is then computed as $D_r = \frac{1}{2} \rho C_D \abs{V}^2 A$. Afterwards, the acceleration (based on the world frame in Figure \ref{fig:worldcoordinate}) is determined in Equation \ref{acceleqn}. Note that $\phi$ \& $\theta$ can be calculated as $\tan^{-1} \frac{U_{object}}{W_{object}}$ \& $\tan^{-1} \frac{V_{object}}{U_{object}}$ respectively.

\begin{equation}
    \label{cd}
    \begin{split}
     C_D = \frac{24}{Re} + \frac{2.6(\frac{Re}{5.0})}{1+(\frac{Re}{5.0})^{1.52}} + \frac{0.411(\frac{Re}{2.63\times10^5})^{-7.94}}{1 +  (\frac{Re}{2.63\times10^5})^{-8.00}} \\ + \frac{0.25(\frac{Re}{10^6})}{1 + (\frac{Re}{10^6})} 
    \end{split}
\end{equation}

\begin{subequations}
\label{acceleqn}
\begin{align}
a_x = - \frac{D_r}{m} \sin{\phi} \cos{\theta}\\
a_y = g - \frac{D_r}{m} \sin{\phi} \sin{\theta}\\
a_z = - \frac{D_r}{m} \cos{\phi} \cos{\theta}
\end{align}
\end{subequations}

\subsubsection*{Step 3: Update of position \& velocity}

The final step updates the position and velocity of the object for the next time step (in the future) using a pre-defined time step size parameter $t_{step}$. This update is done using the fundamental kinematic equations in Equation \ref{kinematiceqn} for the $i$th axis.

\begin{subequations}
\label{kinematiceqn}
\begin{align}
i_{object} = i_{object} + i_{object} t_{step} + \frac{1}{2} a_i t_{step}^2\\
i_{object} = i_{object} + a_i t_{step}
\end{align}
\end{subequations}

\subsubsection*{Completing the Algorithm}

The three steps are repeated for each incremented $t_{step}$ for as long as necessary. A pseudo-code of the algorithm is summarised in Algorithm 1. The path\_list is used in the final stage of the architecture.

\begin{algorithm}
	\caption{Object Trajectory Prediction Pseudo-Code} 
	\begin{algorithmic}[1]
	    \label{predalgo}
        \While{Camera Detects the ball}
            \State \textbf{Initialise} path\_list to store position and time of predicted path
            \State \textbf{Initialise} the time step, time\_step to specified value and the current time, time
            \State Subscribe and obtain $X_{object}, Y_{object}, Z_{object}$ from the Object Detection step.
            \State Apply Step 1 (Equation \ref{linearreg}) to determine $U_{object}, V_{object}, W_{object}$.
            \While{Path is being determined}
                \State Apply Equations \ref{cd} to \ref{acceleqn} in Step 2 to determine $a_x, a_y, a_z$
                \State Apply Equation \ref{kinematiceqn} in Step 3 to update $X_{object}, Y_{object}, Z_{object}$ and $U_{object}, V_{object}, W_{object}$
                \State time $\gets$ time + time\_step
                \State Append $X_{object}, Y_{object}, Z_{object}$ and time to path\_list
                \State Break out of path loop when necessary
            \EndWhile
        \State Publish path\_list for the UAV to use in path planning
        \EndWhile
	\end{algorithmic} 
\end{algorithm}

\subsubsection{Path Planning Scenarios}

The final stage of the proposed architecture explores 3 different methods of path planning that result in 5 scenarios for testing.

\subsubsection*{Method 1 - Cat \& Mouse}

\begin{figure}
    \centering
    \includegraphics[width = 0.5\linewidth]{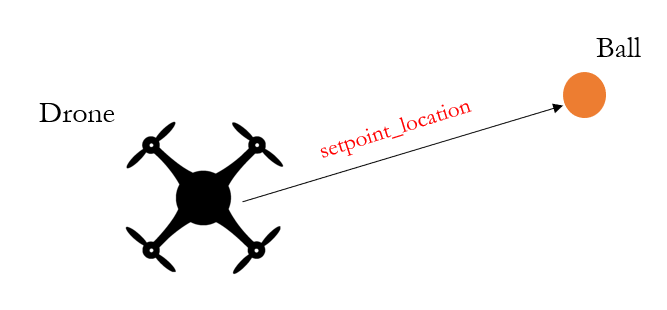}
    \caption{Scenario A: UAV path to stationary ball via the Cat \& Mouse method}
    \label{fig:method1_static}
\end{figure}

This first method that we term "Cat \& Mouse" does not require the predicted path information from Section \ref{trajpred}. In every time interval that the camera detects the object location, the path planner directs the UAV to move to that exact location. Figure \ref{fig:method1_static} illustrates the setpoint command towards a stationary ball. We define this specific scenario as scenario A.

\begin{figure}[H]
    \centering
    \includegraphics[width = 0.7\linewidth]{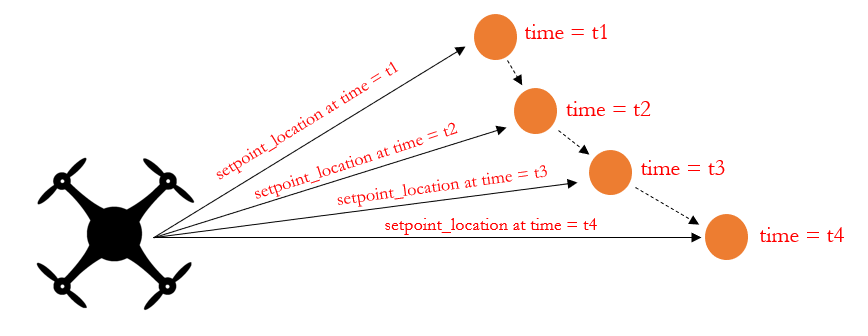}
    \caption{Scenario B: UAV path to moving ball via the Cat \& Mouse method at different time instances}
    \label{fig:method1_moving}
\end{figure}

In the case of a moving object, the UAV will still give commands to the location of the object so long it is in the view of the depth camera. Figure \ref{fig:method1_moving} shows how the setpoint command for the UAV changes at different time intervals while a ball is moving away. We label this as Scenario B.

\begin{figure}[H]
    \centering
    \includegraphics[width =0.8\linewidth]{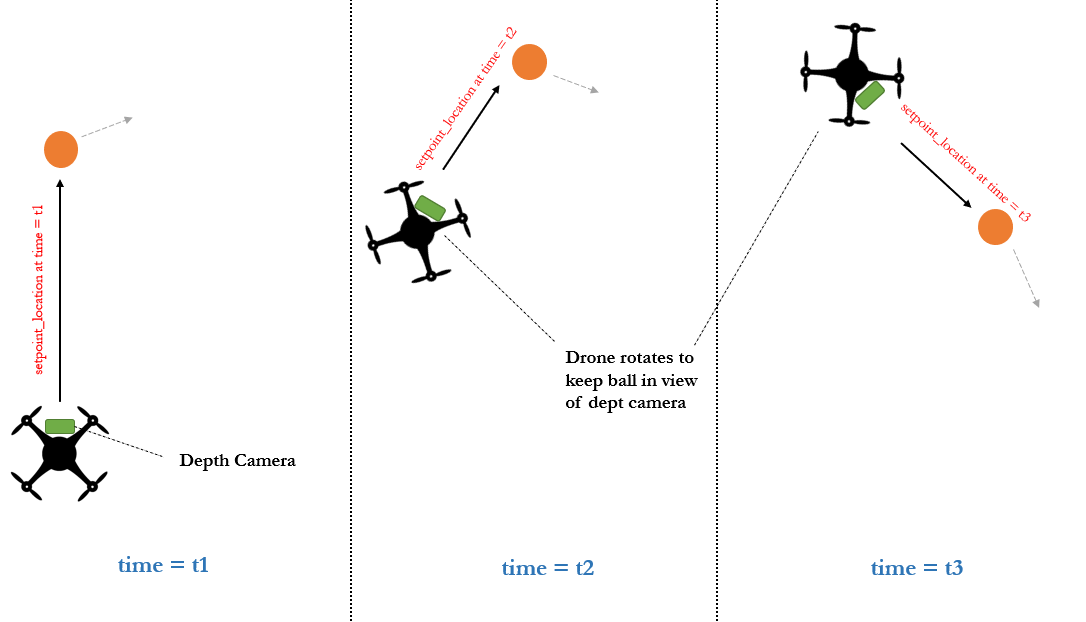}
    \caption{Scenario C: UAV path to moving ball via the Cat \& Mouse method with yaw rotation at different time instances}
    \label{fig:method1_moving2}
\end{figure}

It is possible however for the object to leave the sight of the depth camera, hence the UAV can be programmed to yaw if the detected object appears to be too close to the edge of the camera view based on a threshold. This can also be implemented in this Cat \& Mouse method as illustrated in Figure \ref{fig:method1_moving2}. This illustration resembles the act of a cat chasing a mouse, hence the term was coined as such. We refer to this as Scenario C.

The clear benefit of this method is that it is very easy to implement and computationally inexpensive. Depending on the speed of the depth camera, the path planner generally receives the location of the object at a high frequency. The algorithm for trajectory prediction in Section \ref{trajpred} is computationally demanding and slows down the rate of information transfer to the path planner node.

\begin{figure}[H]
    \centering
    \includegraphics[width =0.6\linewidth]{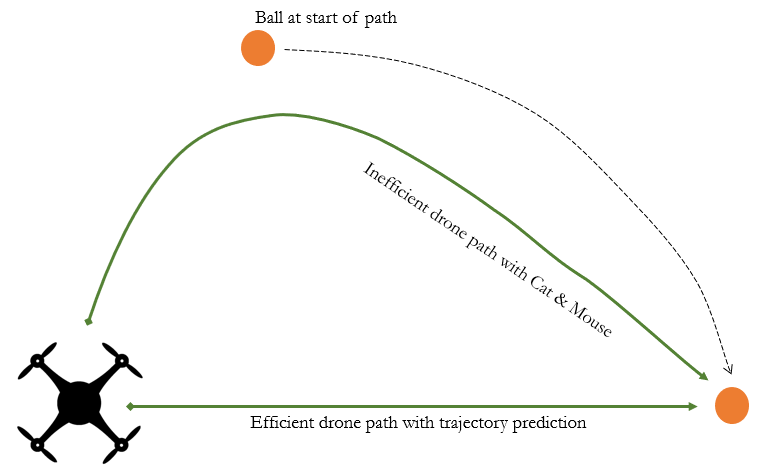}
    \caption{Comparison between a path with and without trajectory prediction}
    \label{fig:effvineff}
\end{figure}

However, the primary issue with this method is that the UAV naturally takes an inefficient path. In the case of a real cat chasing a mouse, the mouse could move in unpredictable ways, so the cat naturally chases directly at the mouse to keep up. However, the trajectory for an object with no external forces is predictable with dynamics and kinematics. The UAV would be more efficient in taking a shorter path to a location where the object would end up instead of following wherever the object goes. Figure \ref{fig:effvineff} illustrates this with the ball.

\subsubsection*{Method 2 - Prediction with Shortest Path}

\begin{figure}[H]
    \centering
    \includegraphics[width =0.6\linewidth]{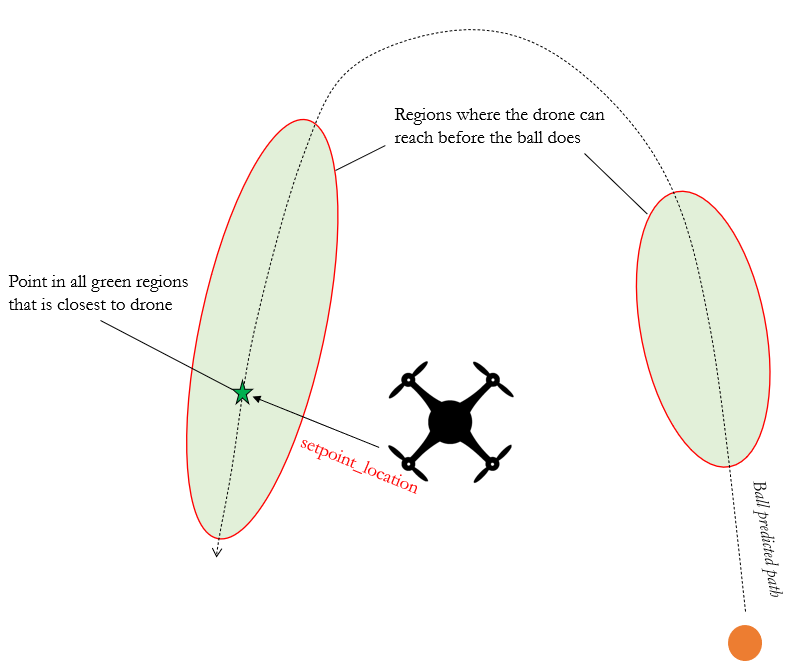}
    \caption{Scenario D: UAV path to a moving ball via the Prediction with Shortest Path method}
    \label{fig:method2}
\end{figure}

Note for methods 2 \& 3, the UAV always yaws to keep the object in the view of the depth camera. The second method, termed "Prediction with Shortest Path" first uses trajectory prediction (as described in Section \ref{trajpred}) to determine the object's path in the future. Note that the trajectory prediction is being continuously run to error correct in-flight as the predictions have errors in real-world conditions. This method then identifies which one of these points along with the point the UAV can reach before the object reaches there. These points form regions as depicted by the green regions in Figure \ref{fig:method2}. The path planner node then selects the point in the green region nearest to the UAV and then directs the setpoint there. We call this Scenario D.

\begin{figure}[H]
    \centering
    \includegraphics[width =0.5\linewidth]{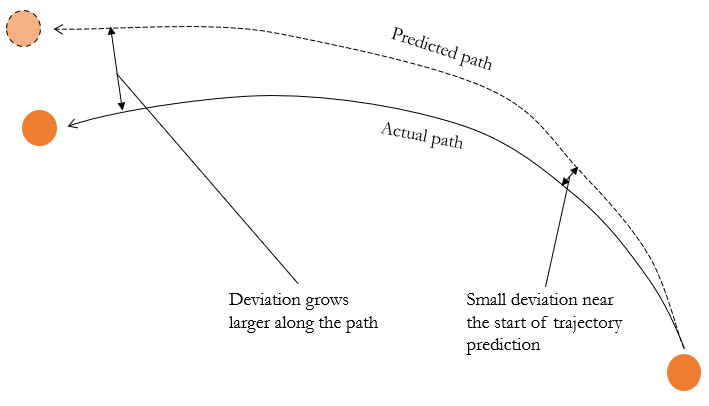}
    \caption{Blowup of small deviation at the start of trajectory prediction along the path}
    \label{fig:pathdev}
\end{figure}

The advantage of this method is that it takes the shortest path, which reduces UAV movement, making the flight fast and efficient. Nonetheless, the drawback is that the UAV may end up moving to a position that is far down the predicted trajectory of the object. Any error of the predicted path of the object close to the object will only get amplified with time, as illustrated in Figure \ref{fig:pathdev}. Depending on the object's path, the closest few points to the UAV are in completely different parts of the predicted trajectory. This may confuse the path planner and it may end up directing completely different setpoints during flight.

\subsubsection*{Method 3 - Prediction with Fastest Path}
\label{method3}

\begin{figure}[H]
    \centering
    \includegraphics[width =0.65\linewidth]{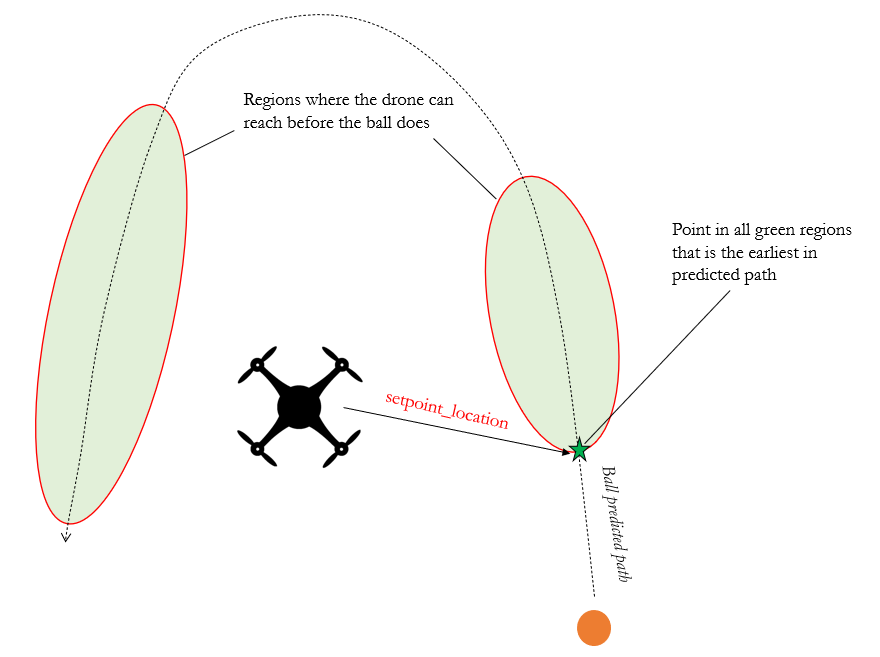}
    \caption{Scenario E: UAV path to a moving ball via the Prediction with Fastest Path method}
    \label{fig:method3}
\end{figure}

In light of the issues faced by the previous method, we have modified it to another method we term "Prediction with Fastest Path". Instead of choosing the closest point, the path planner node chooses the point that is the earliest along the path of the object. This is illustrated in Figure \ref{fig:method3}, where the point in the green region that is earliest in the ball path becomes the setpoint for the UAV. We label this Scenario E. 

This method has some additional benefits. Considering that it approaches the earliest point of the object's path, it is the fastest and most efficient method to intercept the object even if it is not the shortest path. Intercepting the object in the earliest region of the predicted path reduces the predicted error of interception. Nonetheless, it is worth noting that this method relies on very accurate object detection, trajectory prediction and an accurate understanding of its speed. This is very difficult to achieve in a real-world scenario. 

\subsection{Hardware}

Gazebo, an open-source program that offers the ability to simulate robots in any environment, was used to simulate the UAV and the ball intercepting ability. Gazebo provides a physics engine that is highly programmable with a wide variety of open-source models available for use. 


An Iris UAV with a RealSense depth camera attached to it was simulated in Gazebo. MAVROS is used to control the UAV. Once the UAV is armed, it can be controlled by the companion computer. MAVROS takes in data from the path planning node running on the companion computer to control the UAV. The UAV is first sent a setpoint location to hover at an elevation of 2$m$ above the ground. At this point, the UAV is prepped and ready to commence the simulation of our scenarios.

The AMD Ryzen 5-3550H processor along with the NVIDIA GeForce GTX 1050 GDDR5 GPU is used to run simulations for this paper.

\section{Results and Discussion}
Method 1 was used in scenarios A, B and C while methods 2 and 3 were required for the tasks in scenarios D and E. 
\subsection{Scenarios A, B \& C}

\begin{figure}[H]
    \centering
    \includegraphics[width = 0.6\linewidth]{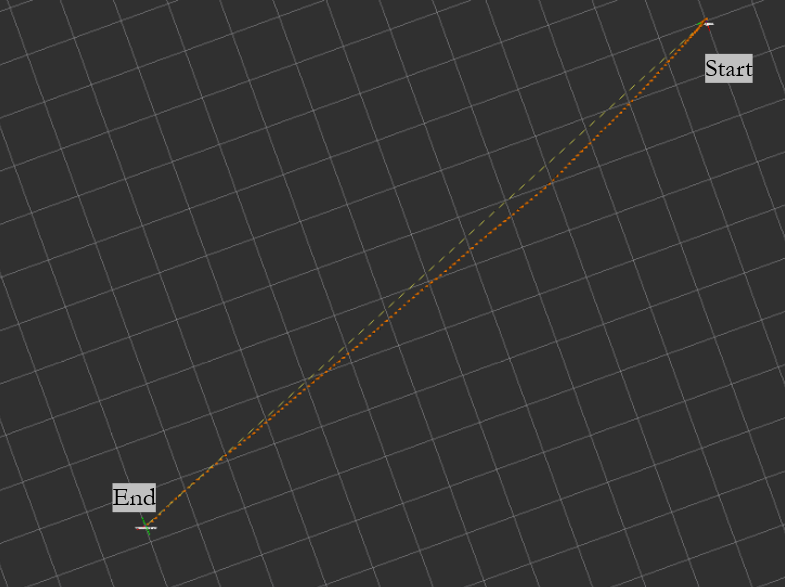}
    \caption{Scenario A: UAV moves directly towards the ball placed directly at the centre of the field of view}
    \label{fig:A1}
\end{figure}

While Scenario A was straightforward, it provided valuable insights that would further more complex scenarios. By tilting during acceleration, the camera often loses sight of the ball. This is detrimental as a continuous view of the ball is necessary for a purely on-board system to work. This can be remedied in two simple ways. First, the angle of pitch is measured through the internal IMU sensor, and the UAV's height would be increased proportionately to keep the ball in view. The other way reduced the UAV's acceleration and the consequent thrust vectoring \& pitch angle. For Scenario A, the latter option was used and the path is shown in Figure \ref{fig:A1}.

\begin{figure}[H]
    \centering
    \includegraphics[width = 0.6\linewidth]{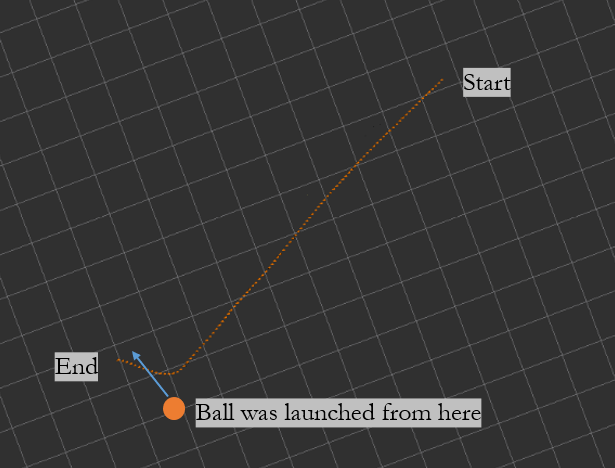}
    \caption{Scenario B: UAV moves towards the moving ball. It takes a sharp turn near the end to chase after the ball but is unable to intercept.}
    \label{fig:B1}
\end{figure}

\begin{figure}[H]
    \centering
    \includegraphics[width = 0.6\linewidth]{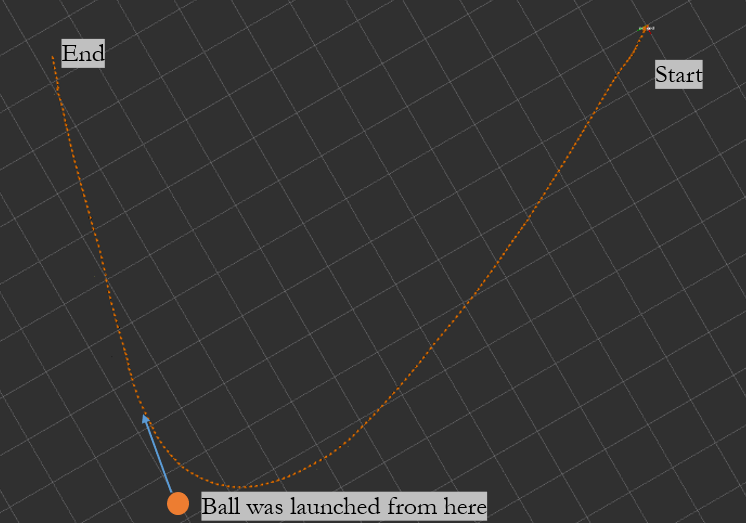}
    \caption{Scenario C: UAV shows a clear curved path chasing after the ball, showing excellent position and yaw control}
    \label{fig:C1}
\end{figure}

The UAV path for Scenario B is shown in Figure \ref{fig:B1}. It was observed that there is an upper limit to the speed of the ball moving relative to the UAV for object detection. This speed limit is tied to the frame rate of the depth camera, as the UAV path planning can only update itself when the ball's new location is updated. It was also noticed that when the UAV moves closer to the ball, the ball will travel past the camera's sensor at a faster rate. At a far enough distance, the UAV would translate to the side, trying to keep the ball in its field of view. However, when it gets close, the UAV's speed is unable to match up to the ball, losing track of the ball as the ball crosses the camera's field of view. This emphasises the need for yawing. Hence it is unsurprising that the UAV and ball paths in Figure \ref{fig:C1} show that Scenario C was met with tremendous success with the UAV reliably intercepting the ball.

\subsection{Scenarios D \& E}

\begin{figure}[H]
    \centering
    \includegraphics[width = 0.8\linewidth]{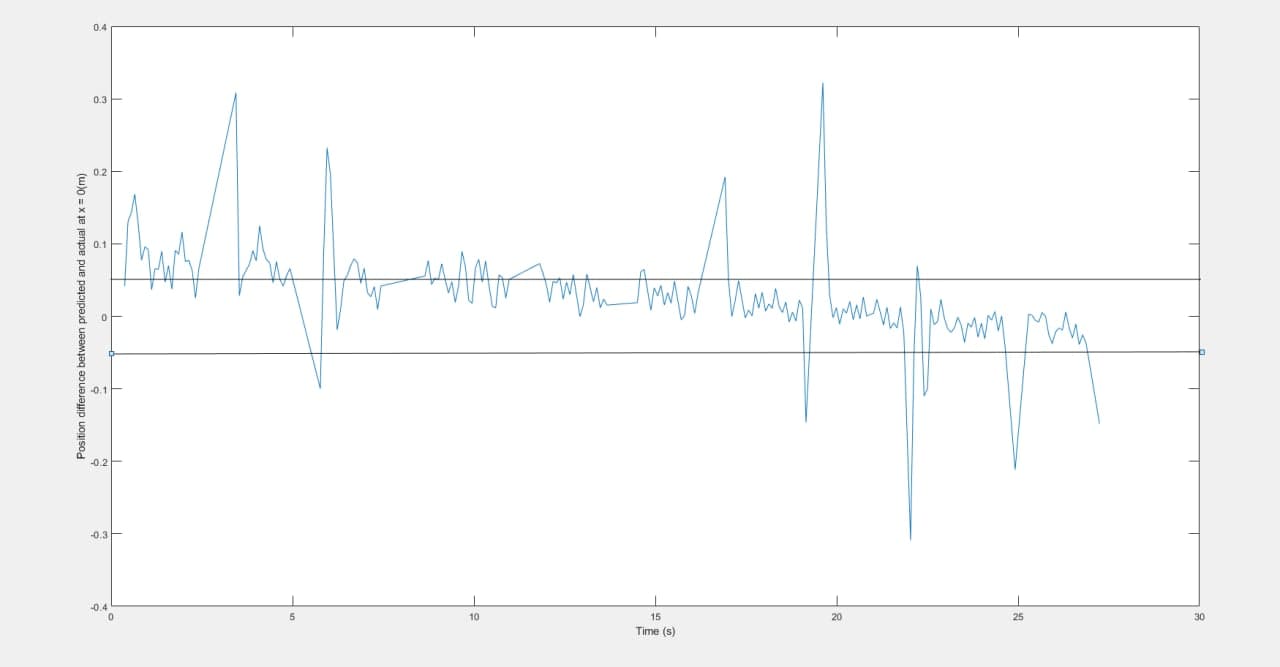}
    \caption{Prediction and Actual position error during ball trajectory for 2D trajectory prediction. Note that the time axis has a 20$\times$ slow-down}
    \label{fig:2d}
\end{figure}

The working demonstrations of scenarios A-C are a prerequisite for trajectory prediction-based methods (2 \& 3) to work. Before simulating these scenarios, a simplified 2D scenario to test trajectory prediction was implemented. In this version, the simulated ball has a trajectory in the general direction of the UAV (but not directly). The UAV is restricted to move within a plane only and it should intercept the ball at the location of the point where the ball intersects the predefined plane. Throughout the trajectory of the ball, the UAV made constant calculations of the expected point of crossing the plane. Figure~\ref{fig:2d} shows the error rate between the actual and predicted locations throughout the trajectory of the ball. It can be seen that the UAV can accurately predict the location on the plane within 0.5m. Given the size of the iris UAV, this meant that the UAV was able to intercept the ball every time by preemptively being at the location on the plane before the ball reached it.

\begin{figure}[H]
    \centering
    \includegraphics[width = 0.7\linewidth]{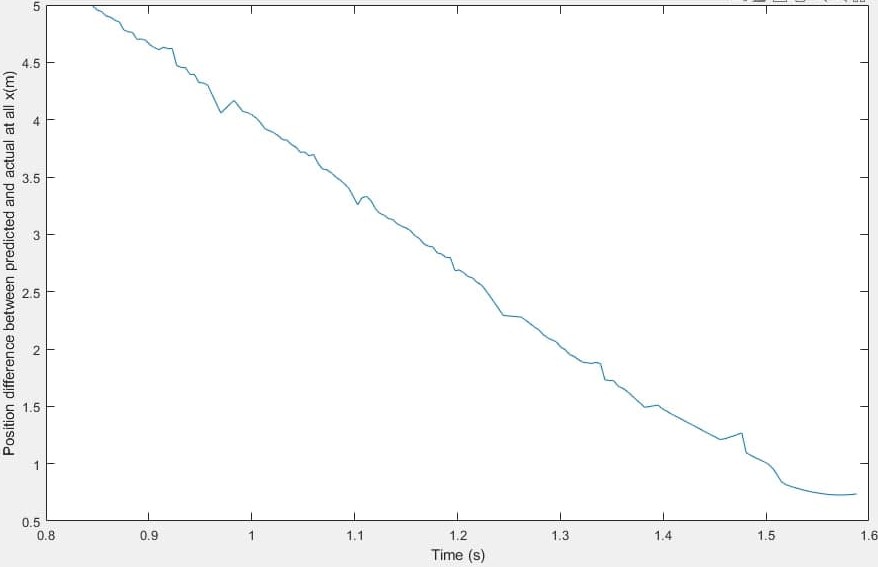}
    \caption{Scenario D: Prediction and Actual position error during flight}
    \label{fig:D}
\end{figure}

Figure~\ref{fig:D} shows the distance between the predicted point of the UAV-ball intersection and its closest point on the real trajectory. The plot shows that the UAV makes extremely inaccurate predictions of the path at the start. This is likely due to the lack of sufficient data points of the ball's location. Throughout the flight of the UAV, the prediction error reduces significantly and the UAV is seen to move to a point where the ball is expected to pass through. Given the large size of the UAV, an error of 0.7m was small enough for the UAV to intercept the ball.
 
\begin{figure}[H]
    \centering
    \includegraphics[width = 0.7\linewidth]{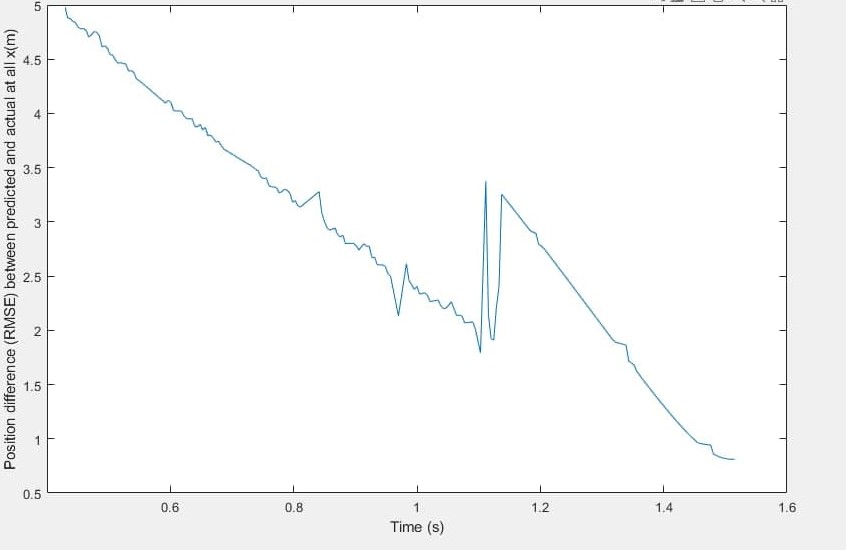}
    \caption{Scenario E: Prediction and Actual position error during flight}
    \label{fig:E}
\end{figure}

Figure~\ref{fig:E} shows a similar plot for Scenario E. Even though the time taken is similar to Scenario D, the path of the ball is different such that the shortest time path is not the same as the shortest distance path. Just like Scenario D, the prediction for Scenario E starts very bad and corrects in-flight to accurately predict the future location of the ball and intercept it. Note that there were few sporadic occasions where the UAV slightly missed the ball. This is an expected downside of the method as described in section~\ref{method3}.

\section{Conclusion}

The present study set out to create a control and software architecture that allows a UAV to detect objects, predict their path and intercept it efficiently without any external aid. Instead, a mounted depth camera is used to automate the process on-board. A variety of path planning methods were introduced along with scenarios to demonstrate a proof-of-concept of the software architecture. All work was done in the Gazebo simulation environment. The results for all scenarios A - E were successful and showed the potential for a seamless hardware and software integration from the depth camera to the UAV path planner. 

The next step in improving the control \& software architecture is to consider self-propelled objects. This is a different challenge as the estimation of an unpredictable object's path makes the trajectory prediction model in the present study not fully applicable. Such a feature would be especially useful in defence applications like a rocket or rogue UAV interception.



\section{Acknowledgment}
The research is supported in part by the Defence Science and Technology Agency (DSTA) - Singapore. 

\bibliographystyle{IEEEtran}
\bibliography{references}

\begin{thebibliography}{10}
\providecommand{\url}[1]{#1}
\csname url@samestyle\endcsname
\providecommand{\newblock}{\relax}
\providecommand{\bibinfo}[2]{#2}
\providecommand{\BIBentrySTDinterwordspacing}{\spaceskip=0pt\relax}
\providecommand{\BIBentryALTinterwordstretchfactor}{4}
\providecommand{\BIBentryALTinterwordspacing}{\spaceskip=\fontdimen2\font plus
\BIBentryALTinterwordstretchfactor\fontdimen3\font minus
  \fontdimen4\font\relax}
\providecommand{\BIBforeignlanguage}[2]{{%
\expandafter\ifx\csname l@#1\endcsname\relax
\typeout{** WARNING: IEEEtran.bst: No hyphenation pattern has been}%
\typeout{** loaded for the language `#1'. Using the pattern for}%
\typeout{** the default language instead.}%
\else
\language=\csname l@#1\endcsname
\fi
#2}}
\providecommand{\BIBdecl}{\relax}
\BIBdecl

\bibitem{Indoor_env}
J.~P. How, B.~Behihke, A.~Frank, D.~Dale, and J.~Vian, ``Real-time indoor
  autonomous vehicle test environment,'' \emph{IEEE Control Systems Magazine},
  vol.~28, no.~2, pp. 51--64, 2008.

\bibitem{precise_aggressive_maneuvers_quadrotors}
D.~Mellinger, N.~Michael, and V.~Kumar, ``Trajectory generation and control for
  precise aggressive maneuvers with quadrotors,'' \emph{The International
  Journal of Robotics Research}, vol.~31, no.~5, pp. 664--674, Jan. 2012.

\bibitem{quad_flips}
S.~Lupashin, A.~Schöllig, M.~Sherback, and R.~D'Andrea, ``A simple learning
  strategy for high-speed quadrocopter multi-flips,'' in \emph{2010 IEEE
  International Conference on Robotics and Automation}, 2010, pp. 1642--1648.

\bibitem{Quad_Ball_Juggle}
M.~Müller, S.~Lupashin, and R.~D'Andrea, ``Quadrocopter ball juggling,'' in
  \emph{2011 IEEE/RSJ International Conference on Intelligent Robots and
  Systems}, 2011, pp. 5113--5120.

\bibitem{Hoffmann2007_line}
G.~Hoffmann, H.~Huang, S.~Waslander, and C.~Tomlin, ``Quadrotor helicopter
  flight dynamics and control: Theory and experiment,'' in \emph{{AIAA}
  Guidance, Navigation and Control Conference and Exhibit}.\hskip 1em plus
  0.5em minus 0.4em\relax American Institute of Aeronautics and Astronautics,
  Jun. 2007.

\bibitem{Cowling2007_polynomial}
I.~D. Cowling, O.~A. Yakimenko, J.~F. Whidborne, and A.~K. Cooke, ``A prototype
  of an autonomous controller for a quadrotor {UAV},'' in \emph{2007 European
  Control Conference ({ECC})}.\hskip 1em plus 0.5em minus 0.4em\relax {IEEE},
  Jul. 2007.

\bibitem{Bouktir_Spline}
Y.~Bouktir, M.~Haddad, and T.~Chettibi, ``Trajectory planning for a quadrotor
  helicopter,'' 07 2008, pp. 1258 -- 1263.

\bibitem{Quadrocopters_Real-Time_Trajectory_Generation}
\BIBentryALTinterwordspacing
M.~Hehn and R.~D{\textquotesingle}Andrea, ``Real-time trajectory generation for
  quadrocopters,'' \emph{{IEEE} Transactions on Robotics}, vol.~31, no.~4, pp.
  877--892, Aug. 2015. [Online]. Available:
  \url{https://doi.org/10.1109/tro.2015.2432611}
\BIBentrySTDinterwordspacing

\bibitem{PID_linearControl}
S.~Bouabdallah, A.~Noth, and R.~Siegwart, ``Pid vs lq control techniques
  applied to an indoor micro quadrotor,'' in \emph{2004 IEEE/RSJ International
  Conference on Intelligent Robots and Systems (IROS) (IEEE Cat.
  No.04CH37566)}, vol.~3, 2004, pp. 2451--2456 vol.3.

\bibitem{augugliaro2013dance}
F.~Augugliaro, A.~P. Schoellig, and R.~D'Andrea, ``Dance of the flying
  machines: Methods for designing and executing an aerial dance choreography,''
  \emph{IEEE Robotics \& Automation Magazine}, vol.~20, no.~4, pp. 96--104,
  2013.

\bibitem{Minimum_snap_trajectory_generation}
D.~Mellinger and V.~Kumar, ``Minimum snap trajectory generation and control for
  quadrotors,'' in \emph{2011 IEEE International Conference on Robotics and
  Automation}, 2011, pp. 2520--2525.

\bibitem{Polynomial_Trajectory_Differential}
\BIBentryALTinterwordspacing
C.~Richter, A.~Bry, and N.~Roy, ``Polynomial trajectory planning for aggressive
  quadrotor flight in dense indoor environments,'' in \emph{Springer Tracts in
  Advanced Robotics}.\hskip 1em plus 0.5em minus 0.4em\relax Springer
  International Publishing, 2016, pp. 649--666. [Online]. Available:
  \url{https://doi.org/10.1007/978-3-319-28872-7_37}
\BIBentrySTDinterwordspacing

\bibitem{LBMPC}
P.~Bouffard, A.~Aswani, and C.~Tomlin, ``Learning-based model predictive
  control on a quadrotor: Onboard implementation and experimental results,'' in
  \emph{2012 IEEE International Conference on Robotics and Automation}, 2012,
  pp. 279--284.

\bibitem{time-optimal_Pontryagin}
\BIBentryALTinterwordspacing
M.~Hehn, R.~Ritz, and R.~D'Andrea, ``Performance benchmarking of quadrotor
  systems using time-optimal control,'' \emph{Autonomous Robots}, vol.~33, no.
  1-2, pp. 69--88, Mar. 2012. [Online]. Available:
  \url{https://doi.org/10.1007/s10514-012-9282-3}
\BIBentrySTDinterwordspacing

\bibitem{Time-optimal_quadrotor_flight}
W.~Van~Loock, G.~Pipeleers, and J.~Swevers, ``Time-optimal quadrotor flight,''
  in \emph{2013 European Control Conference (ECC)}, 2013, pp. 1788--1792.

\bibitem{farhadi2018yolov3}
A.~Farhadi and J.~Redmon, ``Yolov3: An incremental improvement,''
  \emph{Computer Vision and Pattern Recognition, cite as}, 2018.

\bibitem{he2016deep}
K.~He, X.~Zhang, S.~Ren, and J.~Sun, ``Deep residual learning for image
  recognition,'' in \emph{Proceedings of the IEEE conference on computer vision
  and pattern recognition}, 2016, pp. 770--778.

\bibitem{Rusu_ICRA2011_PCL}
R.~B. Rusu and S.~Cousins, ``{3D is here: Point Cloud Library (PCL)},'' in
  \emph{{IEEE International Conference on Robotics and Automation (ICRA)}},
  Shanghai, China, May 9-13 2011.

\bibitem{morrison2013data}
F.~A. Morrison, ``Data correlation for drag coefficient for sphere,''
  \emph{Department of Chemical Engineering, Michigan Technological University,
  Houghton, MI}, vol. 49931, 2013.

\bibitem{chhabra1999drag}
R.~Chhabra, L.~Agarwal, and N.~K. Sinha, ``Drag on non-spherical particles: an
  evaluation of available methods,'' \emph{Powder technology}, vol. 101, no.~3,
  pp. 288--295, 1999.

\end{thebibliography}

\end{document}